\documentclass{kbsjrnl}
\usepackage{harvard}

\begin{document}

\journame{Data Mining and Knowledge Discovery}
\volnumber{vv}
\issuenumber{ii}
\issuemonth{mm}
\volyear{2001}
\received{}\revised{}

\authorrunninghead{Ron Kohavi and Foster Provost}
\titlerunninghead{Data Mining for Electronic Commerce}

\newcommand{\about}{\symbol{126}} 
\newcommand{\rem}[1]{\marginpar{\scriptsize $\leftarrow$ \raggedright #1}}

\setcounter{secnumdepth}{0}

\begin{article}
\bibliographystyle{agsm}

\title{Applications of Data Mining\\ to Electronic Commerce}

\author{Ron Kohavi}
\email{ronnyk@bluemartini.com} 
\affil{Blue Martini Software\\ 
      2600 Campus Dr., 
      San Mateo, CA 94403}
\author{Foster Provost}
\email{provost@acm.org}
\affil{New York University\\
44 W. 4th St., New York, NY 10012}


\noindent Electronic commerce is emerging as the killer domain for
data mining technology.  Is there support for such a bold statement?
Data mining techologies have been around for decades, without moving
significantly beyond the domain of computer scientists, statisticians,
and hard-core business analysts.  Why are electronic commerce systems
any different from other data mining applications?

In his book Crossing the Chasm \cite{Moore:95}, Moore writes, ``There
were too many obstacles to its adoption \ldots inability to integrate it
easily into existing systems, no established design methodologies, and
lack of people trained in how to implement it \ldots'' (p. 23).  What was
``it''?  Artificial intelligence technology, as a product.  Data
mining shares many traits with AI technologies in general, so we
should be concerned that they do not share the same business
fate.\footnote{It should be kept in mind that although the record of
success of AI products has been spotty, AI technologies have seen
remarkable success behind the scenes.}

Notwithstanding several notable successes, data mining projects remain
in the realm of research: high potential reward, accompanied by high
risk.  The risk stems from several sources.  It has been reported by
many \cite{langley-simon-ml-app,process-eighty-twenty}, and has been
our experience, that the ``data mining,'' or algorithmic modeling
phase of the knowledge discovery process occupies at most 20\% of the
effort in a data mining project.  Unfortunately, the other 80\%
contains several substantial hurdles that without heroic effort may
block the successful completion of the project.

The following are five desiderata for success.  Seldom are they they
all present in one data mining application.

\begin{enumerate}

\item Data with rich descriptions.  For example, wide customer
records with many potentially useful fields allow data mining algorithms
to search beyond obvious correlations.

\item A large volume of data.  The large model spaces corresponding to 
rich data demand many training instances to build reliable models.

\item Controlled and reliable data collection.  Manual data entry
and integration from legacy systems both are notoriously problematic;
fully automated collection is considerably better.

\item The ability to evaluate results.  Substantial, demonstrable
return on investment can be very convincing.

\item Ease of integration with existing processes.  Even if pilot
studies show potential benefit, deploying automated solutions to
previously manual processes is rife with pitfalls.  Building
a system to take advantage of the mined knowledge can be a substantial
undertaking.  Furthermore, one often must deal with social and
political issues involved in the automation of a previously manual
business process.

\end{enumerate}

So, why is electronic commerce different?  In short, many of
the hurdles are significantly lower.  Consider those mentioned above.
As compared to ancient or shielded legacy systems, data collection can
be controlled to a larger extent.  We now have the opportunity to
design systems that collect data for the purposes of data mining,
rather than having to struggle with translating and mining data
collected for other purposes.  Data are collected electronically,
rather than manually, so less noise is introduced from manual
processing.  Electronic commerce data are rich, containing information
on prior purchase activity and detailed demographic data.

In addition, some data that previously were very
difficult to collect now are accessible easily.  For example,
electronic commerce systems can record the actions of customers in the
virtual ``store,'' including what they look at, what they put into
their shopping cart and do not buy, and so on.  Previously, in order
to obtain such data companies had to trail customers (in person),
surreptitiously recording their activities, or had to undertake
complicated analyses of in-store videos \cite{paco-why-we-buy}.
It was not cost-effective to collect such data in bulk, and 
correlating them with individual customers is practically impossible.
For electronic commerce systems massive amounts of data can be 
collected inexpensively.\footnote{Gathering such data can be facilitated by
appropriate system design.}

At the other end of the knowledge discovery process sit
implementation and evaluation.  Unlike many data mining applications,
the vehicle for capitalizing on the results of mining---the
electronic commerce system---already is automated.  Therefore the
hurdles of system building are substantially lower, as are the
political and social hurdles involved with automating a manual
process.  Also, because the mined models will fit well with the
existing system, computing return on investment can be much easier.

The lowering of several significant hurdles to the applicability of
data mining will allow many more companies to implement intelligent
systems for electronic commerce.  However, there is an even more
compelling reason why it will succeed.  As implied
above, the volume of data collected by systems for electronic commerce
dwarfs prior collections of commerce data.  Manual analysis will
be impossible, and even traditional semi-automated analyses will
become unwieldy.  Data mining soon will become essential for understanding
customers.

\section{The papers in this special issue}

The mining of electronic commerce data is in its infancy.  The papers in
this special issue give us a peek into the state of the art.  For the
most part, they address the problem of Web merchandizing.

Web merchandising, as distinct for example from marketing, focuses on
how to acquire products and how to make them available.  Electronic
commerce affects the acquisition of products, because (as illustrated
best by Dell Computer Corporation) the supply chain can be integrated
tightly with the customer interface.  Even more intriguing from the
data mining perspective, since customers are interacting with the
computer directly, product assortments, virtual product displays, and
other merchandising interfaces can be modified dynamically, and even
can be personalized to individual customers.

Lawrence, Almasi, Kotlyar, Viveros, and Duri \citeyear{LawrenceEtAl:01}
discuss the application of data mining techniques to supermarket
purchases, in order to provide personalized recommendations.  The
study is based on a project involving IBM and Britain's Safeway
supermarkets, in which customers use palm-top PDAs to compose shopping
lists (based to a large extent on the products they have purchased
previously).  The use of the PDAs increases customer convenience,
because they don't have to walk the aisles for these purchases; they
simply pick them up at the store.  However, it reduces the company's
ability to ``recommend'' products via in-store displays, and the like.

Lawrence et al. go on to show how recommendations can be made instead
on the PDA, using a combination of data mining techniques.  The
recommendations were made to actual customers in two field trials.
After incorporating ``interestingness'' knowledge learned from the
first trial, in the second trial (in a different store) the results
were encouraging notwithstanding several application
challenges.\footnote{Customers may not even look at the
recommendation page; there were 30,000 different products, and the
full recommendation method was not implemented for this field trial.}
Specifically, 25\% of orders included something from the
recommendation list, corresponding to a revenue boost of 1.8\%
(respectable as compared to other promotions).  Perhaps more
important, they show that customers are significantly more likely to
choose high-ranked recommendations than low-ranked ones, indicating
that the algorithms are doing well at modeling the likelihood of
purchasing items previously not purchased.  The study shows intuitive
rules and clusters and relative preferences, demonstrating the
potential of data mining for improving understanding of the
business---which may be useful even in cases where recommendations are
not implemented (or are not effective).

The results of data mining seldom can be used ``out of the box,''
without the involvement of expert users.  Often this is because a
business is reluctant to have unverified models determining important
business decisions.  Just as often, however, the involvment of expert
users is to separate out the few precious nuggets of useful knowledge.
One might ask, isn't this the task data mining is supposed to be
solving?  It is; however, there are different kinds of mining.
Today's tools are rather like strip mining than like the lone
prospector carving out single nuggets of pure gold.  Data mining
algorithms often produce a mass of patterns, much smaller than
the original mountain of data, but still in need of post-processing.

Creating individual consumer profiles for personalized recommendation
(or for other purposes, such as providing dynamic content or tailored
advertizing) exacerbates this problem, because now one may be
searching for patterns individually for each of millions of consumers.
Adomavicius and Tuzhilin \citeyear{AdomaviciusTuzhilin:01} address this
problem.  They show how to automate, partially, the process of
expert-driven validation and filtering of large sets of rules.  Their
method comprises various operators for browsing, grouping, validating,
and filtering rules.  They demonstrate the method by applying it to
data on consumer purchases of beverages---about 2000 households over a
year period.  For example, association-rule mining produced over one
million rules from these data.  In about an hour and a half,
comprising mostly browsing and thinking, the expert-filtering process
had rejected definitively 96\% of the rules, and had used 27,000 rules
to build individual profiles for the households (averaging about 14
rules per profile).

As we've mentioned, electronic commerce systems allow unprecedented
flexibility in merchandizing.  However, flexibility is not a benefit
unless one knows how to map the many options to different situations.
For example, how should different product assortments or merchandising
cues be chosen?  Lee, Podlaseck, Schonberg, and Hoch \citeyear{LeeEtAl:01}
focus on the analysis and evaluation of web merchandising.
Specifically, they analyze the ``clickstreams,'' the series of links
followed by customers on a site.  Their thesis is that the
effectiveness of many on-line merchandising tactics can be analyzed by
a combination of specialized metrics and visualization techniques
applied to clickstreams.

Lee et al. provide a detailed case study of the analysis of
clickstream data from a Web retailer.  The study shows how the
breakdown of clickstreams into subsegments can highlight potential
problems in merchandising.  For example, one product has many
click-throughs but a low click-to-buy rate.  Subsequent analysis shows
that it has a high basket-to-buy rate, but a low click-to-basket rate.
This analysis would allow merchandisers to begin to develop informed
hypotheses about how performance might be improved.  For example,
since this is a high-priced product, one might hypothesize that
customers were lured to the product page and then turned off by the
product's high price.  If this were true, there are several different
actions that might be appropriate (reduce the price, convince the customer
that the product is worth its high price, target the lure better so as not to
``waste clicks,'' etc.).

Spiliopoulou and Pohle \citeyear{SpiliopoulouPohle:01} also study
measuring and improving the success of web sites.  In particular, they
are concerned that success should be evaluated in terms of the
business goal of the web site (e.g., retail sales), and that
treatments should not be limited to measurement alone, but also should
suggest concrete avenues for improvement.  To this end, they discuss
the discovery of navigation patterns, presenting a brief but
comprehensive survey of the state of the art, and also presenting a
method that addresses some of its deficiencies.

They demonstrate their methods on the ``SchulWeb'' site, which
provides information and resources regarding German schools.  They
describe that this site is similar to on-line merchandising sites, but
also that the methods should apply more generally---measuring and
improving success is not limited to sales.  By analyzing sequences,
they observe that users are misusing the search features.  They use
this discovery to improve the interface.  After the change, the
effects are measured; they show an improvement in efficiency.

We close the special issue with a survey of existing ``recommender
systems,'' by Schafer, Konstan, and Riedl \citeyear{SchaferKonstanRiedl:01}.
The degree of use of data mining techniques in such systems can fall
anywhere on the spectrum from trivial (extract a non-personalized,
manually crafted recommendation list) to simple (queries for
straightforward statistics) to complex (collaborative filtering), as
the survey illustrates with a wide variety of real-world electronic
commerce applications that use recommender systems in their day-to-day
operations.  The authors also show that recommender systems are used
for a variety of (business) reasons, and that companies typically use
several different techniques (e.g., they describe seven different
recommendation applications used by Amazon.com).  The different
recommendation tasks include: helping new and infrequent visitors,
building credibility through community, enticing customers to come
back, cross-selling, and building long-term relationships.

Finally, Schafer et al. discuss the challenges that lie ahead for
electronic commerce recommendation applications, from the perspectives of both
research and business---and they include an appendix presenting an
informative analysis of current privacy concerns, which threaten the
continued use of data mining in business and should be taken seriously
by all involved.

\section{A common themes emerging from the papers}

We have argued elsewhere that a significant contribution
applied research papers is highlighting areas that require more
attention from the scientific community \cite{ProvostKohavi:98}.
Reading any of the papers in this special issue, you will find
many examples.  One theme pervades: we need to understand better
how to bring problem-specific knowledge to bear effectively.

Problem-specific knowledge applies throughout the knowledge discovery
process.  For example, one type of knowledge regards useful structure
to the data, which augments the traditional feature-vector
representation.  A common instance of such structure is hierarchies
over data primitives, as are found in product catalogs.  \textit{The
need to be able to incorporate hierarchical background knowledge is
shown in every paper,} with the exception of the survey paper (which
does mention the need to be able to deal with ``rich data'').

We see the need for a variety of other types of background knowledge.
Lawrence et al. discuss that company preference knowledge must be
incorporated---the task is not just to recommend what the customer
will most like, but also what the store would like to sell (popular
new products outside the current shopping pattern, products with high
inventories, products with high profit margins, etc.).  Schafer et
al. discuss that, even from the same data, there are different
fundamental recommendation tasks, also pointing out that there is more
to recommending than just giving the customer what he most would like
to buy.  Really, the system is there to help to improve the
(long-term) business relationship, which has several dimensions.

It also should be kept in mind that there is more to data mining than
just building an automated recommendation system.  If indeed one is
participating in a knowledge discovery process, the knowledge that is
discovered may be used for various purposes.  The papers by Lee et
al. and by Spiliopoulou and Pohle
show knowledge discovery techniques used for understanding the
business more deeply.  Their primary purpose is to shed insight on how
electronic commerce systems might be improved (e.g., by highlighting
problem areas).  
Comprehensibility (beyond data mining
experts) is crucial for successful knowledge discovery, yet we see
relatively little research addressing it \cite{Pazzani:00}.

With the exception of the data mining algorithm, in the current state
of the practice the rest of the knowledge discovery process is manual.
Indeed, the algorithmic phase \textit{is} such a small part of the
process because decades of research have focused on automating it---on
creating effective, efficient data mining algorithms.  However, when
it comes to improving the efficiency of the knowledge discovery
process as a whole, additional research on efficient mining algorithms
will have diminishing returns if the rest of the process remains
difficult and manual.  Adomavicius and Tuzhilin contribute to research
on ``the rest of the process,'' dealing with the often mentioned but
seldom addressed problem of filtering the resultant discoveries.

In sum, the papers in the special issue highlight that although
electronic commerce systems are an ideal application for data 
mining, there still is much research needed---mostly in areas
of the knowledge discovery process other than the algorithmic phase.

\bibliography{editorial}
\end{article}
\end{document}